\documentclass[runningheads]{llncs}
\usepackage{graphicx}
\graphicspath{ {./images/} }
\usepackage{comment}
\usepackage{amsmath,amssymb} 
\usepackage{color}
\usepackage{tabularx}
\usepackage{graphicx}
\usepackage[ruled,vlined]{algorithm2e}
\usepackage{algorithmic}
\usepackage{bm}
\usepackage{hyperref}
\hypersetup{
    colorlinks=true,
    linkcolor=blue,
    filecolor=magenta,     
    urlcolor=blue,
}

\newcommand{\bmh}{\bm{h}}
\newcommand{\bmy}{\bm{y}}
\newcommand{\bmf}{\bm{f}}
\newcommand{\bmz}{\bm{z}}
\newcommand{\bmw}{\bm{w}}
\newcommand{\bmmu}{\bm{\mu}}
\newcommand{\bmsigma}{\bm{\sigma}}
\newcommand{\bmalpha}{\bm{\alpha}}

\begin{document}
\pagestyle{headings}
\mainmatter
\def\ECCVSubNumber{6693}  

\title{Differentiable Joint Pruning and Quantization for Hardware Efficiency}

\titlerunning{Differentiable Joint Pruning and Quantization for Hardware Efficiency}
%
%
\author{Ying Wang\inst{1} \and
Yadong Lu\inst{2}\thanks{Work done during internship at Qualcomm AI Research} \and Tijmen Blankevoort\inst{1}}
\authorrunning{Y. Wang, Y.D. Lu and T. Blankevoort }

\institute{Qualcomm AI Research\thanks{Qualcomm AI Research is an initiative of Qualcomm Technologies, Inc.}\\
\email{\{yinwan, tijmen\}@qti.qualcomm.com} \and
University of California, Irvine\\
\email{yadongl1@uci.edu}}
\maketitle

\begin{abstract}
We present a differentiable joint pruning and quantization (DJPQ) scheme. We frame neural network compression as a joint gradient-based optimization problem, trading off between model pruning and quantization automatically for hardware efficiency. DJPQ incorporates variational information bottleneck based structured pruning and mixed-bit precision quantization into a single differentiable loss function. In contrast to previous works which consider pruning and quantization separately, our method enables users to find the optimal trade-off between both in a single training procedure. To utilize the method for more efficient hardware inference, we extend DJPQ to integrate structured pruning with power-of-two bit-restricted quantization.
We show that DJPQ significantly reduces the number of Bit-Operations (BOPs) for several networks while maintaining the top-1 accuracy of original floating-point models  (e.g.,
53x BOPs reduction in ResNet18 on ImageNet, 43x in MobileNetV2). 
Compared to the conventional two-stage approach, which optimizes pruning and quantization independently, our scheme outperforms in terms of both accuracy and BOPs. Even when considering bit-restricted quantization, DJPQ achieves larger compression ratios and better accuracy than the two-stage approach.


\end{abstract}

\section{Introduction}
There has been an increasing interest in the deep learning community to neural network model compression, driven by the need to deploy large deep neural networks (DNNs) onto resource-constrained mobile or edge devices. So far, pruning and quantization are two of the most successful techniques in compressing a DNN for efficient inference \cite{han2015deep}\cite{he2017channel}\cite{he2018amc}\cite{krishnamoorthi2018quantizing}. Combining the two is often done in practice. However, almost no research has been published in combining the two in a principled manner, even though finding the optimal trade-off between both methods is non-trivial. There are two challenges with the currently proposed approaches:

First, most of the previous works apply a two-stage compression strategy: pruning and quantization are applied independently to the model \cite{han2015deep}\cite{louizos2017bayesian}\cite{ye2018unified}. We argue that the two-stage compression scheme is inefficient since it does not consider the trade-off between sparsity and quantization resolution. For example, if a model is pruned significantly, it is likely that the quantization bit-width has to be large since there is little redundancy left. Further, the heavily pruned model is expected to be more sensitive to input or quantization noise. Since different layers have different sensitivity to pruning and quantization, it is challenging to optimize these holistically, and manually chosen heuristics are probably not optimal.

Second, the joint optimization of pruning and quantization should take into account practical hardware constraints. Although unstructured pruning is more likely to result in higher sparsity, structured pruning is often preferable as it can be more easily exploited on general-purpose devices. Also, typical quantization methods use a fixed bit-width for all layers of a neural network. However, since several hardware platforms can efficiently support mixed-bit precision, joint optimization could automatically support learning the bit-width. The caveat is that only power-of-two bit-widths are often efficiently implemented and supported in typical digital hardware; other bit-widths are rounded up to the nearest power-of-two-two values, resulting in inefficiencies in computation and storage~\cite{ignatov2019ai}. Arbitrary bit-widths used in some literature~\cite{DQ}\cite{HAQ} are more difficult to exploit for power or computational savings unless dedicated silicon or FPGAs are employed.

To address the above challenges, we propose a differentiable joint pruning and quantization (DJPQ) scheme. It first combines the variational information bottleneck~\cite{dai2018compressing} approach to structured pruning and mixed-bit precision quantization into a single differentiable loss function. Thus, model training or fine-tuning can be done only once for end-to-end model compression. Accordingly, we show that on a  practical surrogate measure Bit-Operations (BOPs) we employ, the complexity of the models is significantly reduced without degradation of  accuracy (e.g.,
53x BOPs reduction in ResNet18 on ImageNet, 43x in MobileNetV2).

The contributions of this work are:
\begin{itemize}
\item We propose a differentiable joint optimization scheme that balances pruning and quantization holistically.
\item We show state-of-the-art BOPs reduction ratio on a diverse set of neural networks (i.e., VGG, ResNet, and MobileNet families), outperforming the two-stage approach with independent pruning and quantization. 
\item The joint scheme is fully end-to-end and requires training only once, reducing the efforts for iterative training and finetuning.
\item We extend the DJPQ scheme to the bit restricted case for improved hardware efficiency with little extra overhead. The proposed scheme can learn mixed-precision for a power-of-two bit restriction.
\end{itemize}

\section{Related work}
Both quantization and compression are often considered separately to optimize neural networks, and only a few papers remark on the  combination of the two. We can thus relate our work to three lines of papers: mixed-precision quantization, structured pruning, and joint optimization of the two.

Quantization approaches can be summarized into two categories: fixed-bit and mixed-precision quantization. Most of the existing works fall into the first category, which set the same bit-width for all layers beforehand. Many works in this category suffer from lower compression ratio. For works using fixed 4 or 8-bit quantization that are hardware friendly such as~\cite{RQ}\cite{gysel2018ristretto}\cite{UNIQ}, either the compression ratio or performance is not as competitive as the mixed-precision 
opponents~\cite{DQ}. In~\cite{li2016ternary} 2-bit quantization is utilized for weights, but the scheme suffers from heavy performance loss even with unquantized activations. 
For mixed-precision quantization, a second-order quantization method is proposed in~\cite{dong2019hawq} and \cite{hawq2}, which automatically selects quantization bits based on the largest eigenvalue of the Hessian of each layer and the trace respectively. In~\cite{HAQ}\cite{yazdanbakhsh2018releq}  reinforcement-based schemes are proposed to learn the bit-width. Their methods determine the quantization policy by taking the hardware accelerator’s feedback into the design loop. In~\cite{louizos2017bayesian}, the bit-width is chosen to be proportional to the total variance of that layer, which is heuristic and likely too coarsely estimated. A differentiable quantization (DQ) scheme proposed in~\cite{DQ} can learn the bit-width of both weights and activations. The bit-width is estimated by defining a continuous relaxation of it and using straight-through estimator for the gradients. We employ a similar technique in this paper. For a general overview of quantization, please refer to \cite{nvidiaquant}\cite{krishnamoorthi2018quantizing}.


Structured pruning approaches can also be summarized into two categories: the one with fixed pruning ratio  and with learned pruning ratio for each layer. For the first category, \cite{he2017channel} proposed an iterative two-step channel pruning scheme, which first selects pruned channels with Lasso regression given a target pruning ratio, and then finetunes weights.  In~\cite{li2016pruning} and~\cite{he2018soft}, the pruned channels are determined with $L_1$ and $L_2$ norm,  respectively for each layer under the same pruning ratio. The approach proposed in~\cite{molchanov2016pruning} is based on first-order Taylor expansion for the loss function. To explore inter-channel dependency, many works explore second-order Taylor expansion, such as~\cite{theis2018faster}\cite{peng2019collaborative}.  For the second category, pruning ratio for each layer is jointly optimized across all the layers. Sparsity learned during training with $L_{0}$~\cite{L0} regularization has been explored by several works. In~\cite{he2018amc} the pruning ratio for each layer is learned through reinforcement learning.  The  pruning approach in~\cite{dai2018compressing} relates redundancy of channels to the variational information bottleneck (VIB)~\cite{info-bottle}, adding gates to the network and employing a suitable regularization term to train for sparsity.
As mentioned in~\cite{info-bottle}, the VIBNet scheme has a certain advantage over Bayesian-type of compression~\cite{louizos2017bayesian}, as the loss function does not require additional parameters to describe the priors. We also found this method to work better in practice. In our DJPQ scheme, we utilize this VIBNet pruning while optimizing for quantization jointly to achieve a flexible trade-off between the two. For an overview of structured pruning methods, please refer to \cite{kuzmin2019taxonomy}.

To jointly optimize pruning and quantization, two research works have been proposed~\cite{ye2018unified}\cite{tung2018clip}. Unfortunately, neither of them support activation quantization. Thus, the resulting networks cannot be deployed efficiently on edge hardware. Recently, an Alternating Direction Method of Multipliers (ADMM)-based approach was proposed in~\cite{yang2019learning}. It  formulates the automated compression problem to a constrained optimization problem and solves it iteratively using ADMM. One key limitation is the lack of support for hardware friendliness. First, it is based on unstructured pruning and thus no performance benefit in typical hardware. Second, they use a non-uniform quantization which has the same issue. In addition, the iterative nature of ADMM imposes excessive training time for end-to-end optimization while our DPJQ method does not, since we train in only a single training pass.

\section{Differentiable joint pruning and quantization}
We propose a novel differentiable joint pruning and quantization (DJPQ) scheme for compressing DNNs. Due to practical hardware limitations, only uniform quantization and structured pruning are considered. We will first discuss the quantization and pruning method, and then introduce the evaluation metric, and finally go over joint optimization details.


\subsection{Quantization with learnable mapping}
It has been observed that weights and activations of pre-trained neural networks typically have bell-shaped distributions with long tails~\cite{han2015deep}. Consequently, uniform quantization is sub-optimal for such distributions.  We use a non-linear function to map any weight input $x$ to $\tilde{x}$. Let $q_s$ and $q_m$ be the minimum and maximum value to be mapped, where $0<q_s<q_m$. Let $t>0$ be the exponent controlling the shape of mapping (c.f., Appendix~\ref{app:quant_nonlinear}). $q_m$ and $t$ are learnable parameters, and $q_s$ is fixed to a small value. The non-linear mapping is defined as
\begin{align}\label{eq:x_tilde}
    \tilde{x}=\text{sign}(x) \cdot \begin{cases}
    0, &|x|< q_s\\
   (|x|-q_s)^t,& q_s\leq |x|\leq q_m\\
    (q_m-q_s)^t,    & |x|>q_m
\end{cases}
\end{align}
For activation quantization, we do not use non-linear mapping, i.e., for any input $x$, $\tilde{x}$ is derived  from~\eqref{eq:x_tilde} with  $t$ fixed to 1.
After mapping, a uniform quantization is applied to $\tilde{x}$. Let $d$ be quantization step-size, and let $x_q$ be the quantized value of $x$. The quantized version  is given by
\begin{align}\label{equ:x_q}
    x_q=\text{sign}(x) \cdot \begin{cases}
    0, &|x|< q_s\\
    d\lfloor \frac{(|x|-q_s)^t}{d} \rceil,& q_s\leq |x|\leq q_m\\
    d\lfloor \frac{(q_m-q_s)^t}{d} \rceil,              & |x|>q_m
\end{cases}
\end{align}
where $\lfloor \cdot \rceil$ is the rounding operation. We note that the quantization grid for weights is symmetric and the bit-width $b$ is given by
\begin{align}\label{equ: b_symm}
    b=\log_2\lceil \frac{(q_m-q_s)^t}{d}+1\rceil+1.
\end{align}
For ReLU activations, we use a symmetric unsigned grid since the resulting values are always non-negative. Here, the activation bit-width $b$ is given by
\begin{align}\label{equ: b_unsymm}
     b=\log_2\lceil \frac{(q_m-q_s)^t}{d}\rceil.
\end{align}


Finally, we use the straight through estimation (STE) method~\cite{bengio2013estimating} for back-propagating gradients through the quantizers, which \cite{DQ} has shown to converge fast. The gradients of the quantizer output with respect to $d$, $q_m$ and $t$ are given by
\begin{align}\label{equ: grad_d}
    \bigtriangledown_d x_q=
    \begin{cases}
    \text{sign}(x)\left(\lfloor \frac{(|x|-q_s)^t}{d} \rceil -\frac{(|x|-q_s)^t}{d}\right), &q_s \leq |x|\leq q_m\\
    \text{sign}(x)\left(\lfloor \frac{(q_m-q_s)^t}{d} \rceil - \frac{(q_m-q_s)^t}{d}\right), & |x|> q_m, \\
    0 & \text{otherwise}\\
    \end{cases}
\end{align}

\begin{align}\label{equ: grad_q_m}
\bigtriangledown_{q_m} x_q=\begin{cases}
    0, & |x|\leq q_m\\
    \text{sign}(x) t (q_{m}-q_s)^{t-1} ,   & \text{otherwise}
\end{cases}
\end{align}

\begin{align}\label{equ: grad_t}
\bigtriangledown_{t} x_q=\begin{cases}
    \text{sign}(x) (|x|-q_s)^{t}\log(|x|-q_s), & q_s\leq |x|\leq q_m\\
    \text{sign}(x) (q_m-q_s)^{t}\log(q_m-q_s), & |x| > q_m \\
    0,   & \text{otherwise}
\end{cases}
\end{align}

\subsection{Structured pruning via VIBNet gates}
Our structured pruning scheme is based on the variational information bottleneck (VIBNet)~\cite{dai2018compressing} approach. Specifically, we add multiplicative Gaussian gates to all channels in a layer and learn a variational posterior of the weight distribution aiming at minimizing the mutual information between current layer and next layer's outputs, while maximizing the mutual information between current layer and network outputs. The learned distributions are then used to determine which channels need to be pruned. Assuming the network has $L$ layers, and that there are $c_l$ output channels in the $l$-th layer, $l=1,\cdots, L$. Let $\bmh_l \in \mathcal{R}^{c_l}$ be the output of l-th layer after activation. Let $\bmy \in \mathcal{Y}^{c_L}$ be the target output or data labels. Let $I(\bm x;\bm y)$ denote the mutual information between $\bm x$ and $\bm y$.  The VIB loss function $\mathcal{L}_{\text{VIB}}$ is defined as
\begin{align}\label{eq:vib_loss}
    \mathcal{L}_{\text{VIB}}=\gamma \sum_{l=1}^L I\big(\bmh_l;\bmh_{l-1}\big)-I\big(\bmh_l;\bmy\big), 
\end{align}
where $\gamma > 0$ is a scaling factor.
It then follows that  
\begin{align}
    \bmh_l=\bmz_l\odot \bmf_l\big(\bmh_{l-1}\big), 
\end{align}
where $\bmz_l=\{z_{l,1}, \cdots, z_{l,c_{l}}\}$ is a vector of gates for the l-th layer. The $\odot$ represents the element-wise multiplication operator. $\bmf_l$ represents $l$-th layer mapping function, i.e., concatenation of a linear or convolutional transformation, batch normalization and some nonlinear activation. Let us assume that $\bmz_l$ follows a Gaussian distribution with mean $\bmmu_l$ and variance $\bmsigma_l^2$, which can be re-parameterized as
\begin{align}
    \bmz_l=\bmmu_l+\bm{\epsilon}_l \odot \bmsigma_l
\end{align}
where $\bm{\epsilon}_l\sim \mathcal{N}(0,I)$. It follows that the prior distribution of $\bmh_l$ conditioned on $\bmh_{l-1}$ is
\begin{align} \label{equ:p_h_i}
    p(\bmh_l|\bmh_{l-1})\sim \mathcal{N}\big(\bmmu_l\odot \bmf_l(\bmh_{l-1}), \text{diag}[\bmsigma_l^2 \odot \bmf_l(\bmh_{l-1})^2]\big)
\end{align}
We further assume that the prior distribution of $\bmh_l$ is also Gaussian
\begin{align}\label{equ:q_h_i}
    q(\bmh_l)\sim \mathcal{N}(0,\text{diag}[\bm{\xi}_l])
\end{align}
where $\bm{\xi}_l$ is a vector of variances chosen to minimize an upper bound of $\mathcal{L}_{\text{VIB}}$ (c.f., \eqref{eq:vib_loss}), as given in~\cite{dai2018compressing}
\begin{equation}\label{eq:vib_upper}
    \tilde{\mathcal{L}}_{\text{VIB}} \triangleq \text{CE}(\bmy,\bmh_L)+\gamma \sum_{l=1}^L\sum_{i=1}^{c_l}\log \left(1+\frac{\mu_{l,i}^2}{\sigma_{l,i}^2}\right).
\end{equation}
In \eqref{eq:vib_upper}, $\text{CE}(\bmy,\bmh_L)$ is the cross-entropy loss. The optimal $\bm{\xi}_l$ is then derived as 
\begin{align*}
    \bm{\xi}_l^*=(\bmmu_l^2+\bmsigma_l^2)\mathbf{E}_{\bmh_{i-1}\sim p(\bmh_{i-1})}[\bmf_i(\bmh_{i-1})^2].
\end{align*}
Next, we give the condition for pruning a channel. Let $\bmalpha_l$ be defined as $\bmalpha_l= \bmmu_l^2 / \bmsigma_l^2$.  If $\alpha_{li}<\alpha_{th}$, where $\alpha_{th}$ is a small pruning threshold, the i-th channel in the l-th layer is pruned. The pruning parameters $\{\bmmu_l, \bmsigma_l\}$ are learned during training to minimize the loss function.

\subsection{Evaluation metrics}\label{sec:metrics}
Actual hardware measurement is generally a poor indicator for the usefulness of these methods. There are many possible target-devices, e.g. GPUs, TPUs, several phone-chips, IoT devices, FPGAs, and dedicated silicon, each with their own quirks and trade-offs. Efficiency of networks also greatly depends on specific kernel-implementations, and currently many devices do not have multiple bit-width kernels implemented to even compare on in a live setting. Therefore, we choose a more general metric to measure the compression performance, namely Bit-Operations (BOPs) count, which assumes computations can be done optimally in an ideal world. The BOPs metric has been used by several papers such as~\cite{RQ} and~\cite{bethge2020meliusnet}. Our method can be easily modified to optimize for specific hardware by weighting the contributions of different settings.

We also report results on layerwise pruning ratio and mac counts, for purposes of comparison of the pruning methods. 
Let $p_l$ be the pruning ratio of $l$-th layer output channels. We define $P_l$ to be the layerwise pruning ratio, which is the ratio of the number of weights between the uncompressed and compressed models. It can be derived as 
\begin{align}\label{equ:P_l}
    P_l=1-(1-p_{l-1})(1-p_{l}).
\end{align}
The $l$-th layer Multiply-And-Accumulate (MAC) operations is defined as follows. Let $l$-th layer output feature map have width, height and number of channels of $m_{w,l}$, $m_{h,l}$ and $c_l$, respectively.
Let $k_w$ and $k_h$ be the kernel width and height. The MAC count is computed as 
\begin{align} \label{eq:mac}
    \text{MACs}_l \triangleq (1-p_{l-1})c_{l-1}\cdot (1-p_l)c_l\cdot m_{w,l}\cdot m_{h,l}\cdot  k_w\cdot k_h.
\end{align}
The BOP count in the $l$-th layer $\text{BOPs}_l$ is defined as  
\begin{align}\label{eq:bops}
    \text{BOPs}_l \triangleq \text{MACs}_l \cdot b_{w,l} \cdot b_{a,l-1},
\end{align}
where  $b_{w,l}$ and $b_{a, l}$ denote $l$-th layer weight and activation bit-width.

The BOP compression ratio is defined as the ratio between total BOPs of the uncompressed and compressed models. MAC compression ratio is similarly defined. Without loss of generality, we use the MAC compression ratio to measure only the pruning effect, and use the BOP compression ratio to measure the overall effect from pruning and quantization. As seen from \eqref{eq:bops}, BOP count is a function of both channel pruning ratio $p_l$ and bit-width $b_{w,l}$ and $b_{a, l}$, hence BOP compression ratio is a suitable metric to measure a DNN's overall compression.

\subsection{Joint optimization of pruning and quantization}\label{sec:training}
As described in Section~\ref{sec:metrics}, the BOP count is a function of channel pruning ratio $p_l$ and bit-widths $b_{w,l}$ and $b_{a, l}$. As such, incorporating the BOP count in the loss function allows for joint optimization of pruning and quantization parameters.  When combing the two methods for pruning and quantization, we need to define a loss function that combines both in a sensible fashion. So we define the DJPQ loss function as 
\begin{align}\label{equ:loss}
    \mathcal{L}_{\text{DJPQ}} \triangleq \tilde{\mathcal{L}}_{\text{VIB}} +\beta \sum_{l=1}^L \text{BOPs}_l
\end{align}
 where $\tilde{\mathcal{L}}_{\text{VIB}}$ is the upper bound to the VIB loss given by \eqref{eq:vib_upper}, $\beta$ is a scalar, and $\text{BOPs}_l$ is defined in~(\ref{eq:bops}). 
To compute $\text{BOPs}_l$, $b_{w,l}$ and $b_{a, l}$ are given by \eqref{equ: b_symm} and \eqref{equ: b_unsymm}, respectively, and $b_{a,0}$ denotes DNN's input bit-width. For hard pruning, $p_l$ can be computed as
\begin{align}\label{equ:p_l}
    p_l = \frac{\sum_{i=1}^{c_l}\mathbf{1}{\{\alpha_{l,i}<\alpha_{th}\}}}{c_l}
\end{align}
where $\mathbf{1}\{\cdot \}$ is the indicator function. For soft pruning during training, the indicator function in (\ref{equ:p_l}) is relaxed by a sigmoid function $\sigma(\cdot)$, i.e.,
\begin{align}\label{equ:p_l_soft}
    p_l=\frac{\sum_{i=1}^{c_l}\sigma(\frac{\alpha_{l,i}-\alpha_{th}}{\tau})}{c_l}
\end{align}
where $\tau$ is a temperature parameter. Note that soft pruning makes $p_l$, hence the $\text{BOPs}_l$, differentiable with respect to $\bmmu_l$ and $\bmsigma_l$. 

The parameters of joint optimization of pruning and quantization include $\bmw_l$, $\left(\bmmu_l, \bmsigma_l\right)$ and $\left(q_{m,wl}, d_{w,l}, t_{w,l}, q_{m,al}, d_{a,l}\right)$ for $l=1,\cdots, L$. 
In our experimentation we have found that a proper scaling of the learning rates for different pruning and quantization parameters plays an important role in achieving a good performance. Our experiments also showed that using the ADAM optimizer to automatically adapt the learning rates does not give as good a performance as using the SGD optimizer, provided that suitable scaling factors are set manually. Please refer to Appendix~\ref{app:exp_details} for detailed setting of the scaling factors and other hyper-parameters.
 
\begin{algorithm}[t]
 \textbf{Input}: A neural network with weight $\bmw_l$, $l\in [1, L]$;  training data\\
\textbf{Output}: $b_{w,l}$, $b_{a,l}$, $P_l$, $l\in [1, L]$ \\
\textbf{Parameters}: ${ \bmw_l, \left(\bmmu_l, \bmsigma_l\right), \left(q_{m,wl}, d_{w,l}, t_{w,l}, q_{m,al}, d_{a,l}\right) }$, $l\in [1, L]$ \\
 \textbf{Forward pass}:\\
 Anneal strength $\gamma$ and $\beta$ after each epoch\\
 \For{$l \in \{1, \cdots, L\}$}
 {Draw samples $\bmz_l\sim \mathcal{N}(\bmmu_l,\bmsigma_l^2)$ and multiply to the channel output\\
 Compute  $p_l$ according to  (\ref{equ:p_l_soft})\\
 Compute the layerwise pruning ratio $P_l$ according to (\ref{equ:P_l})\\
 Compute bit-width $b_{w,l}$ and $b_{a,l}$ according to (\ref{equ: b_symm}) and (\ref{equ: b_unsymm})\\
 \If{bit-restricted}
 {Adjust $b_{w,l}$, $d_{w,l}$, $b_{a,l}$ and $d_{a,l}$  according to Algorithm~\ref{alg:adj_bw}}
   Quantize $\bmw_l$ and $\bmh_l$ 
   }
   Compute  $\mathcal{L}_{\text{DJPQ}}$ according to (\ref{equ:loss})\\

 \textbf{Backward pass}:\\
 \For{$l \in  \{1, \cdots, L\}$}
 {Compute the gradients of ($d_{w,l}, d_{a,l}$), ($q_{m,wl},q_{m,al}$),  and $t_{w,l}$  according to  (\ref{equ: grad_d}),  (\ref{equ: grad_q_m}) and (\ref{equ: grad_t}), respectively\\
 Compute the gradients of $\bmmu_l$ and $\bmsigma_l$\\
 Compute the gradients of $\bmw_l$\\
 Update ${ \bmw_l, \left(\bmmu_l, \bmsigma_l\right), \left(q_{m,wl}, d_{w,l}, t_{w,l}, q_{m,al}, d_{a,l}\right) }$ with corresponding learning rate
 }
 \caption{Differentiable joint pruning and quantization}
 \label{alg: JVIB}
\end{algorithm}

\subsection{Power-of-two quantization}\label{sec:power_of_two_quant}
To further align this method with what is feasible in common hardware, in this section we extend DJPQ to the scenario where quantization bit-width $b$ is restricted to power-of-two values, i.e., $b\in \{ 2, 4, 8, 16, 32\}$. As reflected in Alg.~\ref{alg: JVIB}, this extension involves an added step where bit-widths are rounded to the nearest power-of-two representable value, and the quantizer stepsizes are updated accordingly without changing $q_m$, c.f. Alg.~\ref{alg:adj_bw} for details.

\begin{algorithm}[ht]
 \textbf{Input}: stepsize $d_l$, max. range $q_{m,l}$, exponent $t_l$ \\
\textbf{Output}: Adjusted bit-width $b_l'$ and stepsize $d_l'$  \\
Compute $b_l$ according to  (\ref{equ: b_symm}) or (\ref{equ: b_unsymm})\\
 Compute $s_l$: $s_l=\log_2 b_l$\\
 Adjust $s_l$ to $s_l'$: $s_l'=\lfloor s_l\rceil$\\
 Adjust $b_l$ to $b_l'$ with $s_l'$: $b_l'=2^{s_l'}$\\
 Adjust $d_l$ to $d_l'$ with $b_l'$: $d_l'=\frac{(q_m-q_s)^t}{2^{b'-1}-1}$ or $d_l'=\frac{(q_m-q_s)^t}{2^{b'}}$ 
 \caption{Adjust bit-width to power of two integers}
 \label{alg:adj_bw}
\end{algorithm}
This extension only involves a small overhead to the training phase. More specifically, while there is no change in the backward pass, in the forward pass the quantizer bit-width and stepsize are adjusted. The adjustment of the stepsize would result in larger variance in gradients, however, the scheme is still able to converge quickly in practice.

\section{Experiments}
We run experiments on different models for image classification, including VGG7~\cite{li2016ternary} on CIFAR10, ResNet18~\cite{he2016deep} and MobileNetV2~\cite{sandler2018mobilenetv2} on ImageNet. Performance is measured by top-1 accuracy and the BOP count. The proposed DJPQ scheme is always applied to pretrained models. We compare with several other schemes from literature including LSQ~\cite{Esser2020LEARNED}, TWN~\cite{li2016ternary}, RQ~\cite{RQ}, WAGE~\cite{wu2018training} and DQ~\cite{DQ}. We also conduct experiments for the two-stage compression approach, to see how much gain is achieved by co-optimizing pruning and quantization. Furthermore, we conduct experiments of DJPQ with power-of-two quantization, denoted as DJPQ-restrict. We also modify the DQ scheme to do restricted mixed-bit quantization to compare with DJPQ-restrict.  


\subsection{Comparison of DJPQ with quantization only schemes}
\subsubsection{CIFAR10 results}
\begin{figure}[h]
\begin{minipage}[t]{0.5\linewidth}
    \centering
    \includegraphics[width=\textwidth]{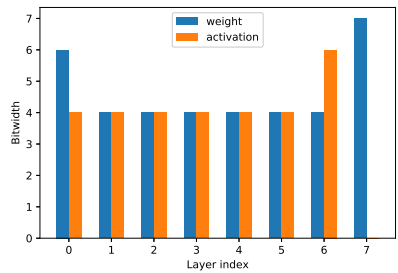}
    \caption{Bit-width of weights and activations for VGG7 with DJPQ scheme}
    \label{fig:bw_vgg7}
    \end{minipage}
\begin{minipage}[t]{0.5\linewidth}
    \centering
    \includegraphics[width=\textwidth]{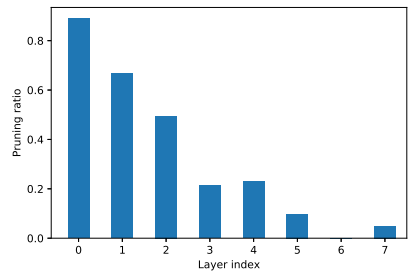}
    \caption{Pruning ratio for VGG7 with DJPQ scheme}
    \label{fig:prune_ratio_vgg7}
    \end{minipage}
\end{figure}

For the VGG7 model on CIFAR10 classification, DJPQ performance along with its baseline floating-point model are provided in Table~\ref{table:vgg7}.  Fig.~\ref{fig:bw_vgg7} shows the weight and activation bit-widths for each layer. The pruning ratio $P_l$ for each layer is given in Fig.~\ref{fig:prune_ratio_vgg7}. Compared to the uncompressed model, we see that DJPQ reduces the amount of BOPs by 210x with less than a 1.5\% accuracy drop. DJPQ is able to achieve a larger compression ratio compared to the other schemes. Comparing e.g. to DQ, which also learns mixed precision, DJPQ has a very similar BOPs reduction.

\setlength{\tabcolsep}{4pt}
\begin{table}
\begin{center}
\caption{VGG7 results on CIFAR10. If weight and activation bit-width are fixed, they are represented  in the format of  weight/activation bit-width in the table. If weight and activation bit-width have different values in different layers, they are denoted as 'mixed'.  Baseline is the floating point model. `BOP comp. ratio' denotes the BOP compression ratio defined in Section~\ref{sec:metrics}. `DJPQ-restrict' denotes DJPQ with power-of-two bit-restricted quantization as presented in Section~\ref{sec:power_of_two_quant}.}
\label{table:vgg7}
\begin{tabular}{lllllll}
\hline
                & bit-width   & Test Acc. &MACs(G) &BOPs(G) & BOP comp. ratio   \\
                \hline
Baseline        & 32/32   & 93.0\%

 & 0.613
   & 629 &--
    \\
    TWN~\cite{li2016ternary} &2/32 &92.56\%  &0.613 &39.23 &16.03 \\
    RQ~\cite{RQ} &8/8 &93.30\%  &0.613 &39.23 &16.03 \\
    RQ~\cite{RQ} &4/4 &92.04\%  &0.613 &9.81 &64.12 \\
    WAGE~\cite{wu2018training} &2/8 &93.22\%  &0.613 &9.81 &64.12\\
DQ\footnotemark[1]~\cite{DQ} &mixed &91.59\%  &0.613 &3.03 &207.59\\
DQ-restrict~\footnotemark[2]~\cite{DQ} &mixed &91.59\% &0.613 &3.40 &185.00\\
\hline
DJPQ &mixed &\textbf{91.54\%} &0.367 &2.99 &\textbf{210.37} \\
DJPQ-restrict &mixed &\textbf{91.43\%} &0.372 &2.92 &\textbf{215.41} \\
\hline
\end{tabular}
\end{center}
\end{table}

\setlength{\tabcolsep}{1.4pt}

\subsubsection{ImageNet results}
\begin{figure}[ht]
\begin{minipage}[t]{0.5\linewidth}
    \centering
    \includegraphics[width=\textwidth]{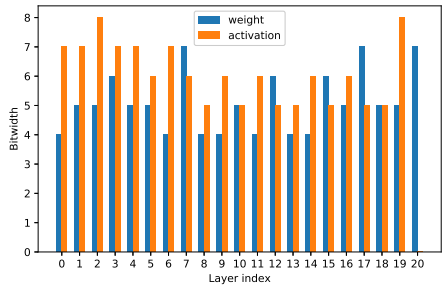}
    \caption{Bit-width  for ResNet18 on ImageNet with DJPQ scheme}
    \label{fig:bw_resnet18}
\end{minipage}
\begin{minipage}[t]{0.5\linewidth}
    \centering
    \includegraphics[width=\textwidth]{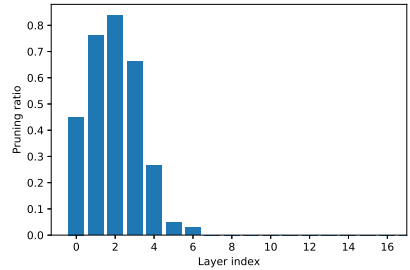}
    \caption{Pruning ratio for ResNet18 on ImageNet with DJPQ scheme}
    \label{fig:prune_ratio_resnet18}
    \end{minipage}
\end{figure}

For experiments on the ImageNet dataset, we applied our DJPQ scheme to ResNet18 and MobileNetV2. Table~\ref{table:resnet18} provides a comparison of DJPQ with state-of-art and other works. The learned bit-width and pruning ratio distributions for DJPQ with 69.27\% accuracy are shown in Figs.~\ref{fig:bw_resnet18} and \ref{fig:prune_ratio_resnet18}, respectively. From the table we see that DJPQ achieves a 53x BOPs reduction while the top-1 accuracy only drops by 0.47\%. The results showed a smooth trade-off achieved by DJPQ between accuracy and BOPs reduction. Compared with other fixed-bit quantization schemes, DJPQ achieves a significantly larger BOPs reduction. Particularly compared with DQ, DJPQ achieves around a 25\% reduction in BOP counts (40.71G vs 30.87G BOPs) with a 0.3\% accuracy improvement. A more detailed comparison of DJPQ and DQ has been given in Appendix~\ref{app:djpq_vs_dq}, which includes weight and activation bit-width distribution, along with pruning ratio distribution. Through the comparison, we show that DJPQ can flexibly trade-off pruning and activation to achieve a larger compression ratio.

\begin{table}[ht]
\centering
\caption{Comparison of ResNet18 compression results on ImageNet. Baseline is the floating point model. `DJPQ-restrict' denotes DJPQ with bit-restricted quantization.}
\label{table:resnet18}
\begin{tabular}{lllllll}
\hline
                & Bit-width   & Test Acc.  &MACs(G) &BOPs(G) & BOP comp. ratio   \\
                \hline
Baseline        & 32/32   & 69.74\%  
 & 1.81   & 1853.44 &--
    \\
LSQ\footnotemark[3]~\cite{Esser2020LEARNED} &3/3 &70.00\%  &1.81 &136.63 &13.56 \\
RQ~\cite{RQ} &8/8 &69.97\%  &1.81 &115.84 &16.00 \\
DFQ~\cite{nagel2019data}  & 4/8   & 69.3\%   & 1.81 &57.92 &32.00       \\
SR+DR~\cite{gysel2018ristretto} &8/8 &68.17\%  &1.81 &115.84 &16.00 \\
UNIQ~\cite{UNIQ} &4/8 &67.02\%  & 1.81 &57.92 &32.00 \\
DFQ~\cite{nagel2019data} &4/4 &65.80\%  &1.81 &28.96 &64.00\\
TWN~\cite{li2016ternary} &2/32 &61.80\%  &1.81 &115.84 &16.00 \\
RQ~\cite{RQ} &4/4 &61.52\%  &1.81 &28.96 &64.00\\
DQ\footnotemark[1]~\cite{DQ} &mixed &68.49\%  &1.81 &40.71 &45.53\\
DQ-restrict~\footnotemark[2]~\cite{DQ} &mixed &68.49\% &1.81 &58.68 &31.59\\
\hline
VIBNet~\cite{dai2018compressing}+fixed quant. &8/8 &69.24\%  &1.36 &87.04 &21.29\\
VIBNet~\cite{dai2018compressing}+DQ\footnotemark[1]~\cite{DQ} &mixed &68.52\% &1.36 &39.83 &46.53\\
\hline
DJPQ &mixed &\textbf{69.27\%}  &1.39 &35.01 &\textbf{52.94} \\
DJPQ &mixed &68.80\%  &1.39 &30.87  &60.04 \\
\hline
DJPQ-restrict &mixed &\textbf{69.12\%} &1.46 &35.45 &\textbf{52.28} \\
\hline
\end{tabular}
\end{table}

\begin{figure}[ht]
\begin{minipage}[t]{0.5\linewidth}
    \centering
    \includegraphics[width=\textwidth]{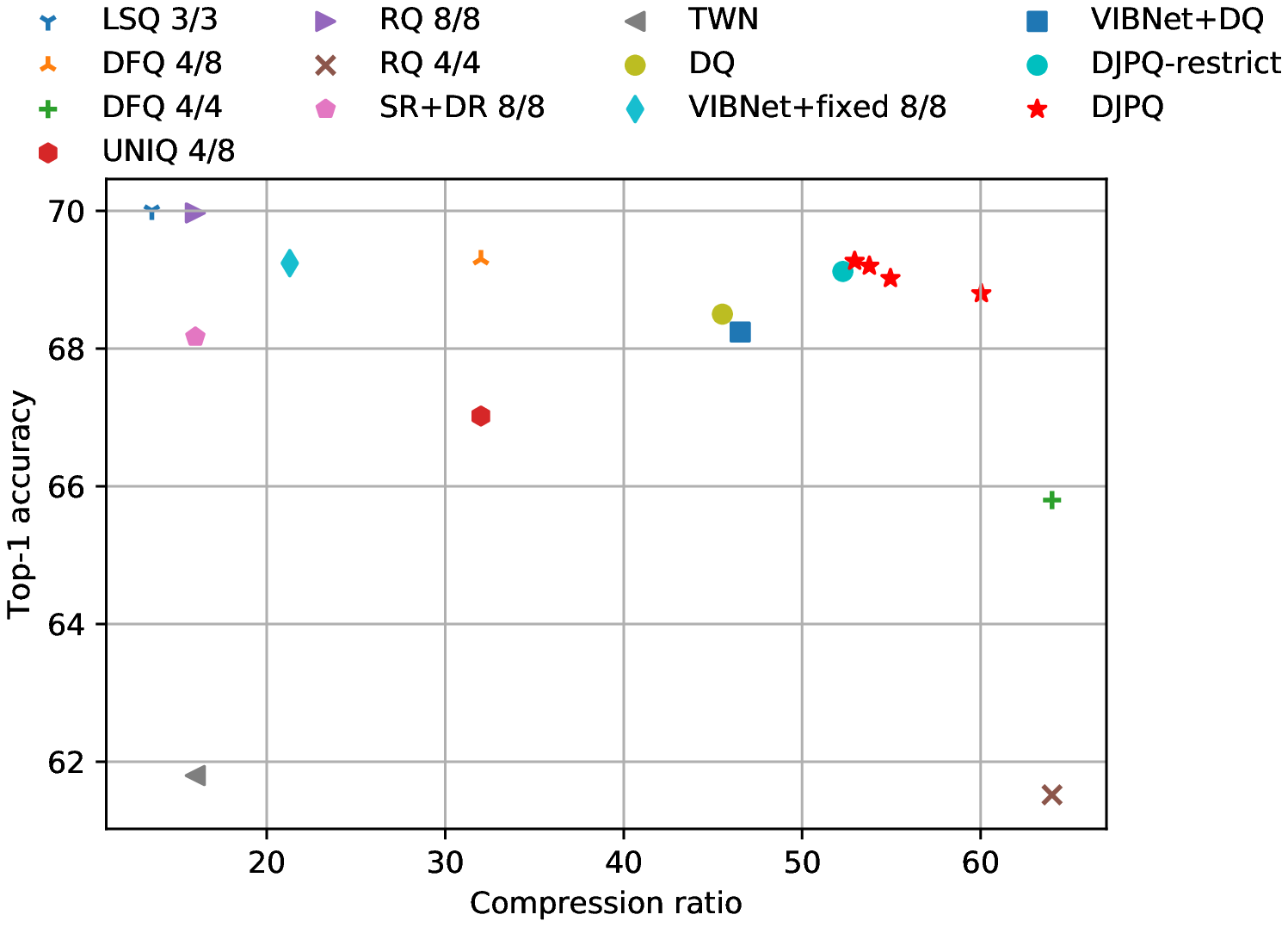}
    \caption{Comparison of BOPs reduction  for ResNet18 on ImageNet}
    \label{fig:bop_resnet18}
\end{minipage}
\begin{minipage}[t]{0.5\linewidth}
    \centering
    \includegraphics[width=\textwidth]{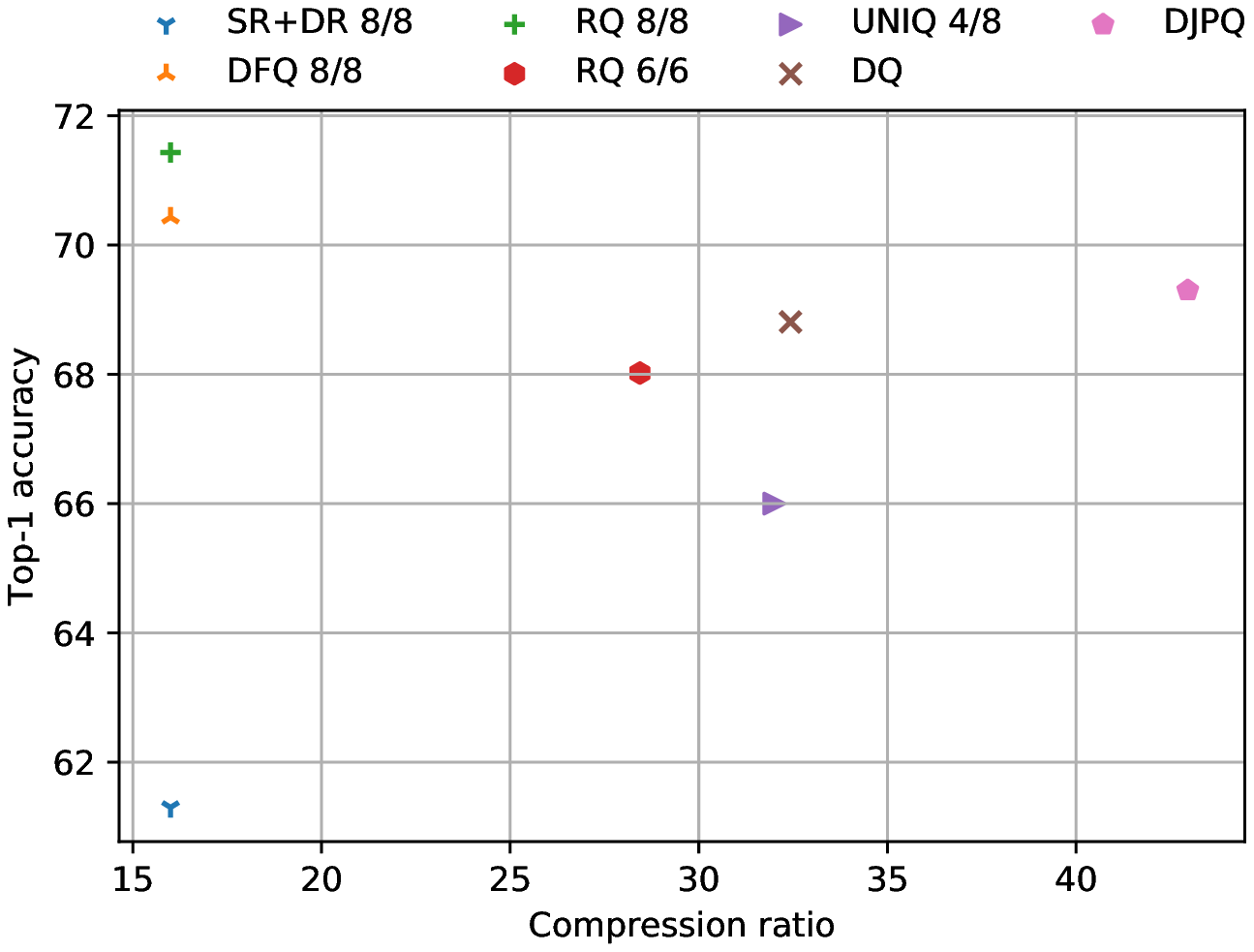}
    \caption{Comparison of BOPs reduction for MobileNetV2 on ImageNet}
    \label{fig:bop_mobilenetv2}
    \end{minipage}
\end{figure}

MobileNetV2 has been shown to be very sensitive to quantization~\cite{nagel2019data}.  Despite this, we show in Table~\ref{table:mobilenetv2} that the DJPQ scheme is able to compress MobileNetV2 with a large compression ratio. The optimized bit-width and pruning ratio distributions are provided in Appendix~\ref{app:mobilenetv2_dist}. It is observed that DJPQ achieves a 43x BOPs reduction within 2.4\% accuracy drop. Compared with DQ, DJPQ achieves around 25\% reduction in BOP counts over DQ on MobileNetV2 (175.24G vs 132.28G BOPs) with 0.5\% accuracy improvement.  Comparisons of BOPs for different schemes are plotted in Figs.~\ref{fig:bop_resnet18} and \ref{fig:bop_mobilenetv2} for ResNet18 and MobileNetV2, respectively. DJPQ provides superior results for ResNet18 and MobileNetV2, and can be easily extended  to other architectures.

\setlength{\tabcolsep}{4pt}
\begin{table}[ht]
\centering
\caption{Comparison of MobileNetV2 compression results on ImageNet.  Baseline is the uncompressed floating point model.}
\label{table:mobilenetv2}
\begin{tabular}{lllllll}
\hline
                & Bit-width   & Test Acc.  &MACs(G) &BOPs(G) &BOP comp. ratio  \\
                \hline
Baseline        & 32/32   & 71.72\%  &5.55 &5682.32 &-- 
    \\
SR+DR~\cite{gysel2018ristretto} &8/8 &61.30\%  &5.55 &355.14 &16.00\\
DFQ~\cite{nagel2019data}  & 8/8   & 70.43\%   &5.55 &355.14 &16.00\\
RQ~\cite{RQ} &8/8 &71.43\%   &5.55 &355.14 &16.00 \\
RQ~\cite{RQ} &6/6 &68.02\%  &5.55 &199.80 &28.44\\
UNIQ~\cite{UNIQ} &4/8 &66.00\%  &5.55 &177.57 &32.00 \\
DQ\footnotemark[1]~\cite{DQ} &mixed &68.81\%  &4.81 &175.24 &32.43\\
\hline
DJPQ &mixed &\textbf{69.30\%}  &4.76 &132.28 &\textbf{42.96} \\
\hline
\end{tabular}
\end{table}

\footnotetext[1]{For a fair comparison, we replaced the memory regularization term in DQ~\cite{DQ} with BOPs regularization. }
\footnotetext[2]{DQ-restrict~\cite{DQ} refers to the scheme where bit-width learned by DQ are upper rounded to the nearest power-of-two integers. }
\footnotetext[3]{LSQ~\cite{nagel2019data} does not quantize the first and last layers.}


\subsection{Comparison of DJPQ with two-stage optimization}
In this section we provide a comparison of DJPQ and the two-stage approach - pruning first, then quantizing, and show that DJPQ outperforms  both in BOPs reduction and accuracy. To ensure a fair comparison, pruning and quantization are optimized independently in the two-stage approach. We have done extensive experiments to find a pruning ratio that results in high accuracy. The two-stage results in Table~\ref{table:resnet18} are derived by first pruning a model with VIBNet gates to a 1.33x MAC compression ratio. The accuracy of the pruned model is 69.54\%. Then the pruned model is quantized with both fixed-bit and mixed-precision quantization. For fixed 8-bit quantization, the two-stage scheme obtains a 21.29x BOP reduction with 69.24\% accuracy. For the VIBNet+DQ approach, we achieve a 46.53x BOP reduction with 68.52\% accuracy. 

By comparing the resulting MAC counts of the compressed model of DJPQ with the two-stage approach, we see a very close MAC compression ratio from pruning between the two schemes (1.24x vs 1.33x). They also have comparable BOP counts, however, DJPQ achieves 0.75\% higher accuracy (69.27\% vs 68.52\%) than the two-stage scheme. The results provide good evidence that even under similar pruning ratio and BOP reduction, DJPQ is able to achieve a  higher accuracy. This is likely due to the pruned channel distribution of DJPQ being dynamically adapted and optimized jointly with quantization resulting in a higher accuracy.

\subsection{Comparison of DJPQ with others under bit restriction}
We further run experiments of the DJPQ scheme for power-of-two bit-restricted quantization. For VGG7 on CIFAR10, the DJPQ result with bit-restricted quantization is provided in Table~\ref{table:vgg7} named `DJPQ-restrict'. With DJPQ compression, VGG7 is compressed by 215x with a 91.43\% accuracy. The degradation of accuracy compared to DJPQ without bit restriction is negligible, showing that our scheme also works well for power-of-two restricted bit-widths.
For ResNet18 on ImageNet, Table~\ref{table:resnet18} shows DJPQ-restrict performance. 
DJPQ-restrict is able to compress ResNet18 by 53x with only a 0.5\% accuracy drop. We also provide a comparison of DJPQ and DQ both with bit restrictions, denoted as `DJPQ-restrict' and `DQ-restrict', respectively. Compared with DQ, DQ-restrict has a much smaller compression ratio. It shows that the performance of compression would be largely degraded if naively rounding the trained quantization bits. DJPQ-restrict has significant compression ratio gain over DQ-restrict results. Comparing DJPQ-restrict to other schemes with fixed 2/4/8-bit quantization,  it is clear that DJPQ-restrict has the highest compression ratio.


\begin{figure}[ht]
\begin{minipage}[t]{0.5\linewidth}
    \centering
    \includegraphics[width=\textwidth]{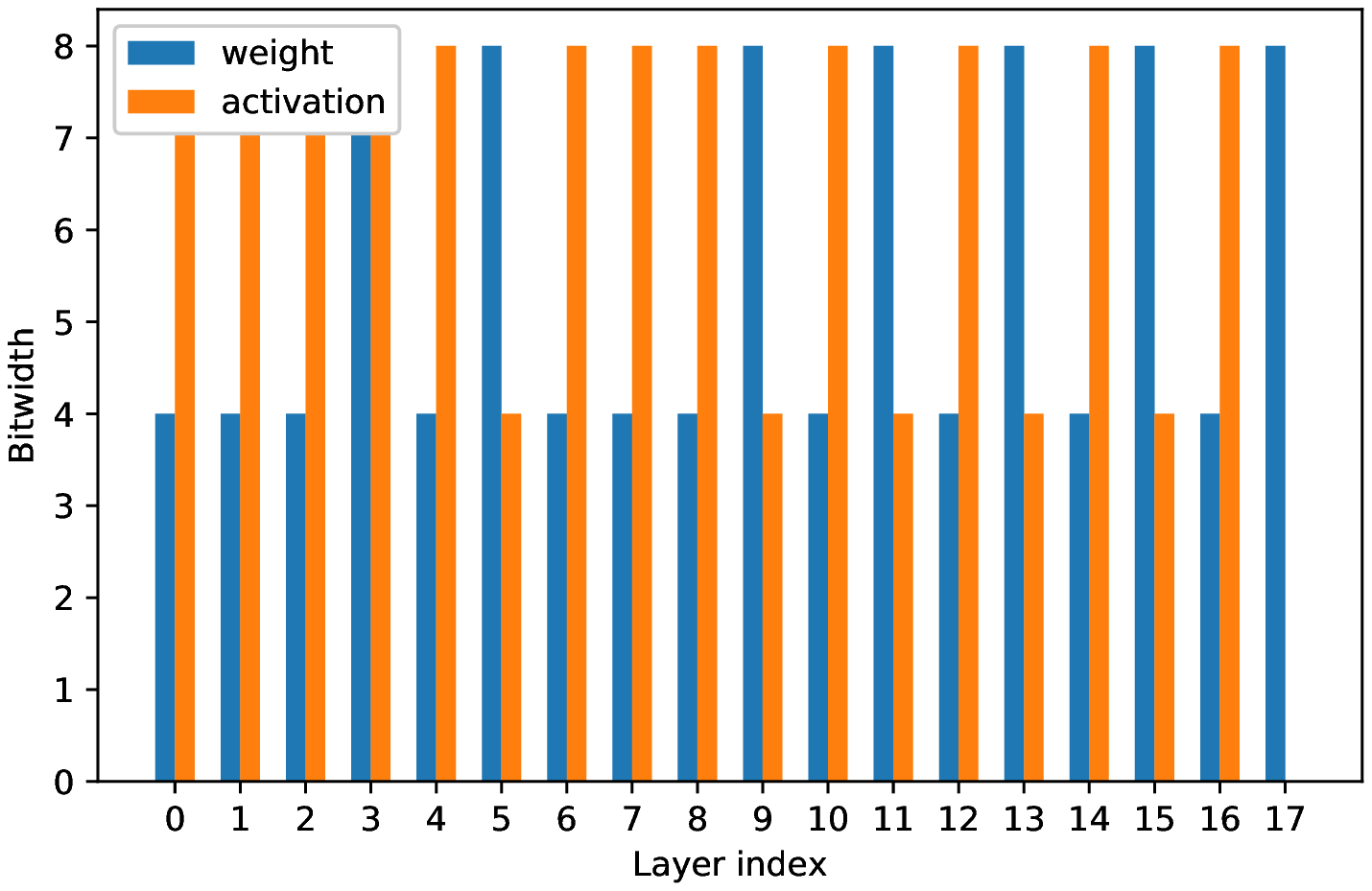}
    \caption{bit-width  for ResNet18 on ImageNet. Bits are restricted to power of 2.}
    \label{fig:bw_resnet18_restd}
\end{minipage}
\begin{minipage}[t]{0.5\linewidth}
    \centering
    \includegraphics[width=\textwidth]{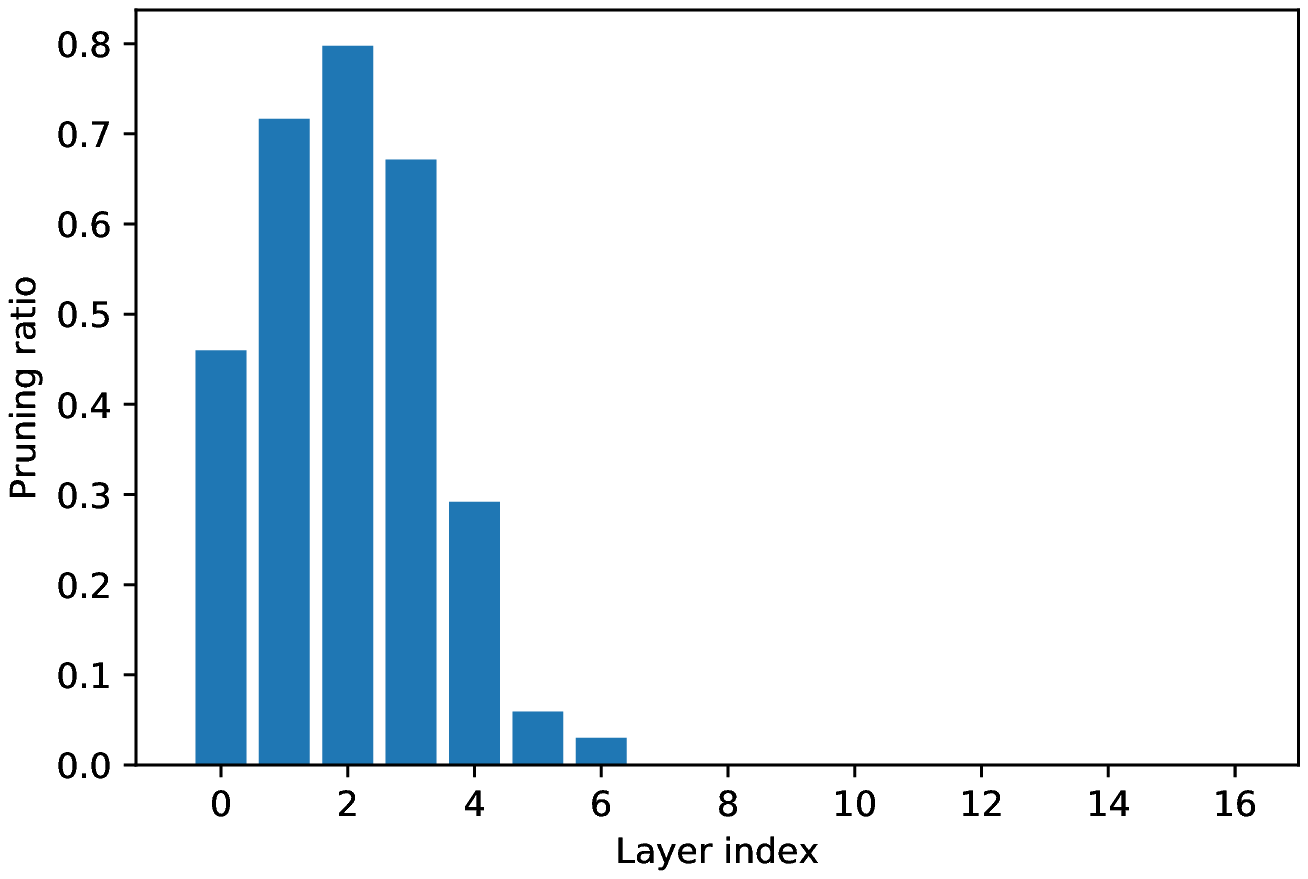}
    \caption{Pruning ratio for ResNet18 on ImageNet. Bits are restricted to power of 2.}
    \label{fig:prune_ratio_resnet18_restd}
    \end{minipage}
\end{figure}


\subsection{Analysis of learned distributions}
For ResNet18, the learned bit-width and pruning ratio distributions for DJPQ with 69.27\% accuracy are shown in Figs.~\ref{fig:bw_resnet18} and \ref{fig:prune_ratio_resnet18}, respectively. It is observed that pruning occurs more frequently at earlier layers. The pruning ratio can be very large, indicating heavy over-parameterization in that layer. Layers \{7,12,17\} are residual connections. From  Fig.~\ref{fig:bw_resnet18} we see that all the three residual connections require larger bits than their corresponding regular branches. Regarding to the distribution of $t$ in the nonlinear mapping, it is observed that for layers with heavy pruning, $t$ is generally smaller and $t<1$; for layers with no pruning, $t$ is close to 1. This gives a good reflection of interaction between pruning and quantization. For MobileNetV2, we found that point-wise convolution layers require larger bit-width than depth-wise ones, indicating that point-wise convolutional layers are more sensitive than depth-wise ones.

\section{Conclusion}
We proposed a differentiable joint pruning and quantization (DJPQ) scheme  that optimizes bit-width and pruning ratio simultaneously. The scheme integrates variational information bottleneck, structured pruning and mixed-precision quantization, achieving a flexible trade-off between sparsity and bit precision. We show that DJPQ is able to achieve larger bit-operations (BOPs) reduction over conventional two-stage compression while maintaining the state-of-art performance. Specifically, DJPQ achieves 53x BOPs reduction in ResNet18 and 43x reduction in MobileNetV2 on ImageNet. We further extend DJPQ to support power-of-two bit-restricted quantization with a small overhead. The extended scheme is able to reduce BOPs  by 52x on ResNet18 with almost no accuracy loss.  

\subsection*{Acknowledgments}
We would like to thank Jilei Hou and Joseph Soriaga for consistent support, and thank Jinwon Lee, Kambiz Azarian  and Nojun Kwak for their great help in revising this paper and providing valuable feedback.

\clearpage
%
%
\bibliographystyle{splncs04}
\bibliography{egbib}
\clearpage
\appendix

\section{Quantization scheme in DJPQ}\label{app:quant_nonlinear}
\begin{figure}[h]
    \centering
    \includegraphics[width=.6\columnwidth]{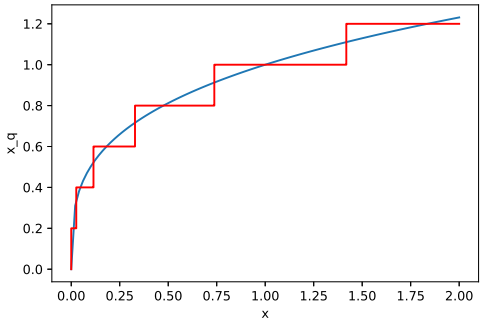}
    \caption{Illustration of quantization scheme. The blue curve gives the nonlinear mapping function. The red curve corresponds to the  quantization value.}
    \label{fig:quant_mapping}
\end{figure}

Fig.~\ref{fig:quant_mapping} illustrates the proposed quantization scheme. First, a non-linear function is applied to map any weight input $x$ to $\tilde{x}$ (shown in blue curve). Then a uniform quantization is applied to $\tilde{x}$. The quantized value $x_q$ is shown in red.
\section{Experimental details}
\subsection{Comparison of DJPQ with DQ}\label{app:djpq_vs_dq}
\begin{figure}[h]
    \centering
    \includegraphics[width=.8\columnwidth]{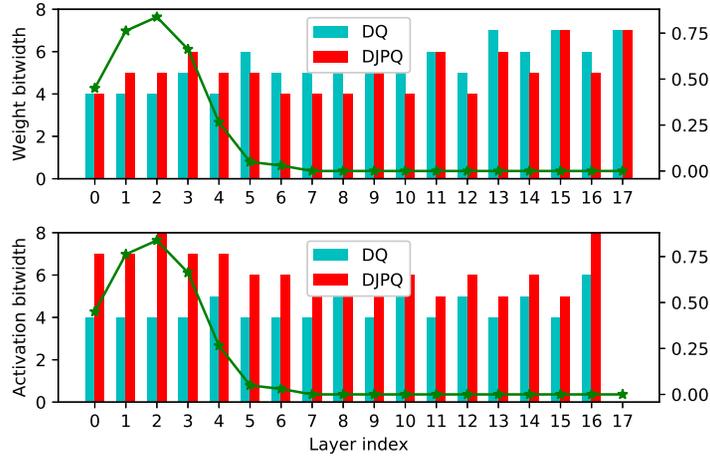}
    \caption{Comparison of bit-width distributions of DQ and DJPQ for ResNet18 on ImageNet. The top figure plots the weight bit-width for each layer, while the bottom plots the corresponding activation bit-width. The green curve is the pruning ratio of DJPQ in each layer .}
    \label{fig:bw_dist_dq_vs_jpq}
\end{figure}

To show that joint optimization of pruning and quantization outperforms quantization only scheme such as DQ, we plot in Fig.~\ref{fig:bw_dist_dq_vs_jpq} a comparison of weight and activation bit-width for DQ and DJPQ. The results are for ResNet18 on ImageNet. We plot the pruning ratio curve of DJPQ in both figures to better show the pruning effect in the joint optimization scheme. 
As seen in the figure,  there is no big difference between weight bit-width for DQ and DJPQ. However, the difference between activation bit-width for the two schemes is significant. Layer 0 to 6 in DJPQ has much larger activation bit-width than those in DQ, while those layers correspond to high pruning ratios in DJPQ. It provides a clear evidence that pruning and quantization can well tradeoff between each other in DJPQ,  resulting in lower redundancy in the compressed model.

\subsection{DJPQ results for MobileNetV2}\label{app:mobilenetv2_dist}
\begin{figure}[h]
    \centering
    \includegraphics[width=\columnwidth]{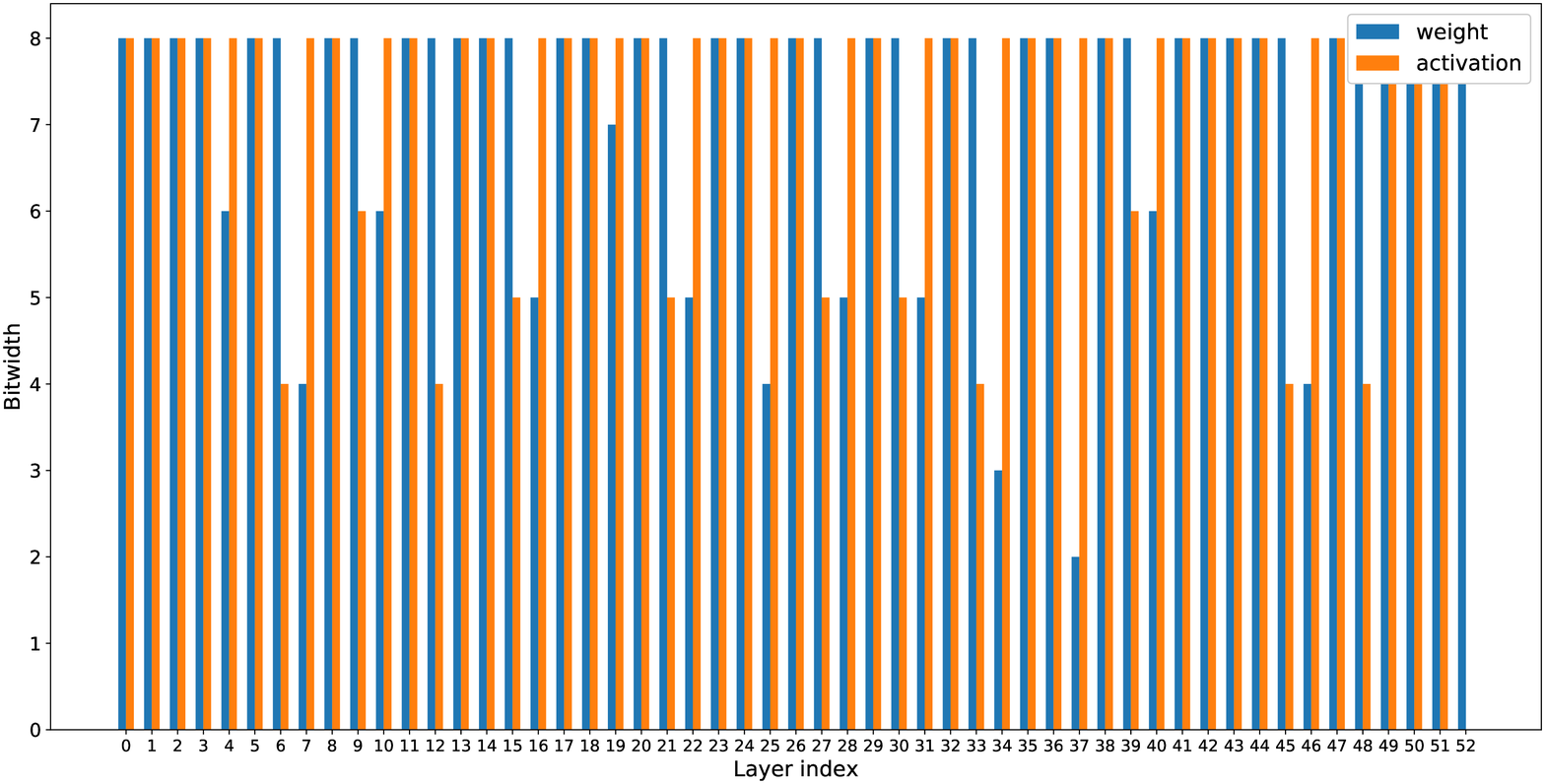}
    \caption{Bit-width for MobileNetV2 on ImageNet with DJPQ scheme}
    \label{fig:bw_mobilenetv2}
\end{figure}

\begin{figure}[ht]
    \centering
    \includegraphics[width=.6\columnwidth]{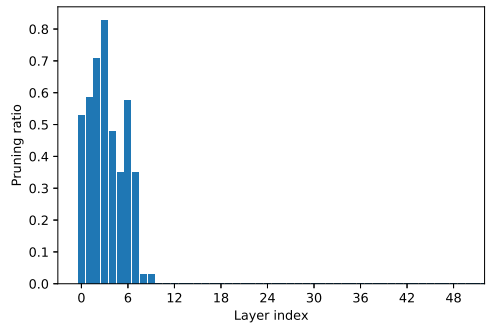}
    \caption{Pruning ratio for MobileNetV2 on ImageNet with DJPQ scheme}
    \label{fig:prune_ratio_mobilenetv2}
\end{figure}

Figs.~\ref{fig:bw_mobilenetv2} and \ref{fig:prune_ratio_mobilenetv2} show the optimized bit-width and pruning ratio distributions, respectively. It is observed that earlier layers tend to have larger pruning ratios than the following layers. And for many layers in MobileNetV2, DJPQ is able to quantize the layers into to small bit-width.

\subsection{Experimental setup}\label{app:exp_details}
The input for all experiments are uniformly quantized to 8 bits.  For the two-stage scheme including `VIBNet+fixed quant.' and `VIBNet+DQ', VIBNet pruning is firstly optimized at learning rate 1e-3 with SGD, with a pruning learning rate scaling of 5. The strength $\gamma$ is set to 5e-6. In DQ for the pruned model, $\beta$ is chosen to 1e-11 and the learning rate is 5e-4. The learning rate scaling for quantization is 0.05. All the pruning threshold $\alpha_{th}$ is chosen to 1e-3. The number of epochs is 20 for each of the stage.

For DJPQ experiments on VGG7,  the learning rate is set to 1e-3 with an ADAM optimizer. The initial bit-width is 6. The strength $\gamma$ and  $\beta$ are 1e-6 and 1e-9, respectively.  The scaling of learning rate for pruning and quantization is 10 and 0.05, respectively. For DJPQ on ResNet18, we chose a learning rate 1e-3 with SGD optimization. The initial bit-width for weights and activations are 6 and 8. The scaling of learning rate for pruning  and quantization are 5 and 0.05, respectively. The strength $\gamma$ and  $\beta$ are 1e-5 and 1e-10.  For DJPQ on MobileNetV2, the learning rate is set to 1e-4 for SGD optimization. The initial bit-width is 8. The strength $\gamma$ and $\beta$ are set to 1e-8 and 1e-11. The learning rate scaling for pruning and quantization are selected to be 1 and 0.005, respectively. 


\end{document}